\documentclass[10pt]{article} 

\usepackage[preprint]{rlj} 

\usepackage{amssymb}            
\usepackage{mathtools}          
\usepackage{mathrsfs}           
\usepackage{graphicx}           
\usepackage{subcaption}         
\usepackage[space]{grffile}     
\usepackage{url}                
\usepackage{lipsum}             

\usepackage{cancel}
\usepackage{bbold}
\usepackage{algpseudocode}
\usepackage{algorithm}
\usepackage{svg}
\usepackage{xspace}

\usepackage{soul}

\newcommand{\dcil}{{\sc dcil}\xspace}

\newcommand{\dcilii}{{\sc dcil-ii}\xspace}

\newcommand{\sac}{{\sc sac}\xspace}
\newcommand{\hac}{{\sc hac}\xspace}
\newcommand{\hrac}{{\sc hrac}\xspace}
\newcommand{\hiro}{{\sc hiro}\xspace}
\newcommand{\tdm}{{\sc tdm}\xspace}
\newcommand{\her}{{\sc her}\xspace}

\newcommand{\tddd}{{\sc td3}\xspace}
\newcommand{\ddpg}{{\sc ddpg}\xspace}

\newcommand{\leap}{{\sc leap}\xspace}

\newcommand{\mymath}[1]{\ensuremath{#1}\xspace}

\newcommand{\taug}{\mymath{\tau_{\mathcal{G}}}}

\newcommand{\tdddher}{{\sc td3+her}\xspace}
\newcommand{\myopictddd}{{\sc td3+her+seq}\xspace}
\newcommand{\mseqtddd}{{\sc $M_{gseq}$-td3}\xspace}
\newcommand{\twogtddd}{{\sc $M_{2g}$-td3}\xspace}

\newcommand{\dubins}{{\em Dubins Hallway}\xspace}
\newcommand{\pointmaze}{{\em PointMaze: Serp3}\xspace}
\newcommand{\cartpole}{{\em GC-Cartpole}\xspace}

\def\bbbe{{\rm I\!E}} 
\newcommand{\Esp}{{\bbbe}{}}

\title{A tale of two goals: leveraging sequentiality in multi-goal scenarios}

\setrunningtitle{Addressing multi-goal scenarios in RL}

\author{Olivier Serris, Stéphane Doncieux, Olivier Sigaud}

\emails{olivier.serris@isir.upmc.fr, stephane.doncieux@isir.upmc.fr, olivier.sigaud@isir.upmc.fr}

\affiliations{
Sorbonne Université, CNRS, ISIR, F-75005, Paris, France
}
\contribution{We show failures when a goal-conditioned policy only aware of the next goal is used to follow sequences of goals.}
{Several hierarchical reinforcement learning methods (\cite{eysenbach_search_2019}, \cite{levy2019learning}) rely on goal-conditioned (GC) policies that are iteratively conditioned on each individual goal to traverse sequences of goals.}

\contribution{We propose two specific instances of Markov Decision Process framework designed to reach several sequences of goals crafted by a fixed planner.}
{Prior work dealt with the case of a single sequence of goals \citep{chenu_leveraging_2023}}

\contribution{In our evaluation environments, we show that targeting the next two goals is more stable and sample efficient than targeting the next and final goals.}
{We train \tddd+\her agents on each MDP formulation, using an expert-crafted planner to supply intermediate goal sequences that lead to final goals. We focus on differences in low-level training, and all policies are evaluated on navigation and pole balancing tasks.}


\summary{Several hierarchical reinforcement learning methods leverage planning to create a graph or sequences of intermediate goals, guiding a lower-level goal-conditioned (GC) policy to reach some final goals. The low-level policy is typically conditioned on the current goal, with the aim of reaching it as quickly as possible. However, this approach can fail when an intermediate goal can be reached in multiple ways, some of which may make it impossible to continue toward subsequent goals. To address this issue, we introduce two instances of Markov Decision Process (MDP) where the optimization objective favors policies that not only reach the current goal but also subsequent ones. In the first, the agent is conditioned on both the current and final goals, while in the second, it is conditioned on the next two goals in the sequence. We conduct a series of experiments on navigation and pole-balancing tasks in which sequences of intermediate goals are given. By evaluating policies trained with \tddd+\her on both the standard GC-MDP and our proposed MDPs, we show that, in most cases, conditioning on the next two goals improves stability and sample efficiency over other approaches.
}

\begin{document}

\makeCover  
\maketitle  

\begin{abstract}
Several hierarchical reinforcement learning methods leverage planning to create a graph or sequences of intermediate goals, guiding a lower-level goal-conditioned (GC) policy to reach some final goals. The low-level policy is typically conditioned on the current goal, with the aim of reaching it as quickly as possible. However, this approach can fail when an intermediate goal can be reached in multiple ways, some of which may make it impossible to continue toward subsequent goals. To address this issue, we introduce two instances of Markov Decision Process (MDP) where the optimization objective favors policies that not only reach the current goal but also subsequent ones. In the first, the agent is conditioned on both the current and final goals, while in the second, it is conditioned on the next two goals in the sequence. We conduct a series of experiments on navigation and pole-balancing tasks in which sequences of intermediate goals are given. By evaluating policies trained with \tddd+\her on both the standard GC-MDP and our proposed MDPs, we show that, in most cases, conditioning on the next two goals improves stability and sample efficiency over other approaches.
\end{abstract}

\section{Introduction}

In reinforcement learning (RL), an agent learns to control a system to accomplish a task over a sequence of actions by maximizing some reward signal. Standard RL methods struggle when the sequence of actions is long and the reward signal is sparse. Goal-conditioned (GC) policies \citep{schaul_universal_2015} are not exceptions, often failing to reach distant goals. To address this, several hierarchical reinforcement learning (HRL) approaches propose to decompose the task into a sequence of intermediate goals, with the GC policy iteratively conditioned on each goal until the final one is reached \citep{eysenbach_search_2019,levy2019learning}.

However, as shown in \cite{chenu_leveraging_2023}, when intermediate goals do not fully specify the states in which they are achieved, a {\em chaining} issue arises: the system may achieve a goal in a state that is incompatible with the achievement of the next goal. To counteract this, \cite{chenu_leveraging_2023} propose the \dcilii framework in which the agent prepares for the next goal while aiming at the current one. Their solution integrates successive goals into the Markov Decision Process (MDP) and conditions the policy on both the next goal and its corresponding goal index, allowing the agent to complete a fixed sequence of goals reliably while also leveraging the powerful reward propagation mechanism provided by Hindsight Experience Replay (\her) \citep{andrychowicz_hindsight_2018}.

In this paper, we address a more challenging context in which the agent must learn to achieve any goal from any starting state, and a high-level planner provides various sequences of intermediate goals depending on the final goal. In this context, learning a GC policy that prepares for the next goal may not be enough, as the way to achieve the next goal may itself depend on the final goal.

Our main contribution is to adapt the \dcil framework for multi-goals tasks. We propose two Markov Decision Processes (MDPs) in which the state, transition and reward functions account for sequential goal reaching and make it possible to benefit from goal relabeling as in \her. In the first MDP, the policy is conditioned on the current and next goals, and in the second one, it is conditioned on the current and final goals.

Then, to evaluate these instances, we implement four agents on top of the same actor-critic architecture: an agent conditioned only on the final goal, a myopic agent that only learns to reach the current goal but is iteratively guided by a sequence of intermediate goals, and our solutions: two agents conditioned on the current goal and, respectively, on the next goal and the final goal. We compare these agents on various benchmarks in which the agent is trained to reach any goal from any state given the corresponding sequences of intermediate goals. Our results show that agents based on our framework overcome limitations of the myopic and non-sequential agents. In particular, we show that conditioning on the next two goals is more efficient than doing so on the next and final goals, as it requires to propagate the value in the critic over a shorter horizon.

\section{Background \& Related Work}

All the agents used in this work are goal-conditioned, they rely on the same actor-critic algorithm, they leverage hindsight relabeling and they can be seen as the low-level part of a hierarchical approach. In this section we provide background information about these elements and present some related algorithms.

\subsection{TD3 and SAC}

The agents we compare in this paper all use the twin delayed deep-deterministic policy gradient (\tddd) actor-critic algorithm
\citep{fujimoto_addressing_2018}. \tddd inherits from the \ddpg algorithm \citep{lillicrap2015continuous}
where the actor and the critic are deterministic neural networks. The main innovation of \tddd with respect to \ddpg consists in using two critic networks and taking the minimum of their Q-values to prevent overestimation bias.

The Soft Actor-Critic (\sac) algorithm \citep{haarnoja_soft_2018} can be seen as an extension of \tddd where the policy is stochastic and both the critic and the policy are trained under an entropy regularized regime.

\subsection{Goal-conditioned reinforcement learning}
\label{sec:gcrl}
Goal conditioned reinforcement learning (GCRL) can be formalized as an extension of Markov Decision Processes (MDPs) that we note
$\mathcal{M}_{gc} = (S_{gc},A,R_{gc},P,\gamma,\mathcal{T})$.
At the beginning of each episode, an initial state $s_0$ from the initial state distribution $\rho_0$ and a single goal $g\in G$ are selected. The goal space $G$ is generally defined as a subset of the state space. For instance, the goal might be only the position of the center of mass of a robot, without considering the information relative to its limbs. At each step, the agent observes $(s_t, g) \in S_{gc}=S\times G$. The agent chooses an action $a_t\in A$, moves to a new state $s_{t+1}$ according to an unknown probability distribution $p(s_{t+1}|s_t,a_t) \in P$ and then receives a reward $r_t$. 

If $S_g$ is the set of all states for which $g$ is achieved, the reward is defined as: 
$R_{gc}(s_t,g,a_t,s_{t+1}) = \begin{cases}
    1 &\text{if } s_{t+1}\in S_{g}, \\
    0 &\text{otherwise.}
\end{cases}$

The discount factor $\gamma$ defines the importance of future rewards. 
Finally, $\mathcal{T}(g)$ defines the set of terminal states based on the current goal $g$. 
Some states may be terminal independently from the current goal (e.g. a robot falling on the floor). Reaching a goal can be terminal or not depending on the task at hand. For instance, in a maze environment, the task ends as soon as the agent reaches the goal position, while in a pole-balancing task, the pole must stay at equilibrium in the goal state until the episode ends.

The objective of GCRL algorithms is to find a policy that maximizes the expected cumulative reward:
\begin{equation}
\Esp_{\substack{a\sim \pi_{\theta} \\ s_0 \sim \rho_0 \\ g \sim P(g) }}\Bigg[\sum_{t=0}^{\infty} \gamma^t R_{gc}(s_t,g,a_t,s_{t+1}) \prod_{i=0}^{t}1-I[s_i\in \mathcal{T}(g)]\Bigg].
\end{equation}

\textbf{Hindsight relabeling}:
Learning directly from collected data is very hard, as an untrained agent rarely or never reaches its goal. This sparse reward regime can completely prevent learning. 
A critical ingredient for performance and sample efficiency is the Hindsight Experience Replay (\her) algorithm \citep{andrychowicz_hindsight_2018}. During actor and critic updates, the agent learns from transitions $\{ s_t,g,a_t,r(s_{t+1},g),s_{t+1}\}$ where most of the time, the agent does not receive a reward.
In the following notation, we show how \her modifies transitions by substituting the current goal with a goal achieved later in the trajectory, where $ag_t$ represents the goal achieved at time step $t$ and $t_{max}$ represents the last step of a given trajectory:
\begin{equation}
(s_t,\hbox{\st{$g,R_{gc}(s_t,g,a,s_{t+1})$}},a_t,s_{t+1}) \rightarrow (s_t,ag_{k},R_{gc}(s_t,ag_{k},a_t,s_{t+1}),a_t,s_{t+1}), \text{where } k\in [t,t_{max}].
\end{equation}

\subsection{Hierarchical RL}
\label{sec:feudal}

When the agent's goal is too far from its current state, combining GCRL with \her is not enough. In that case, one can leverage a divide-and-conquer approach by using hierarchical RL (HRL). 

HRL approaches decompose long horizon RL problem into sequences of shorter problems. Generally speaking, a high-level component or a hierarchy of such components selects a sequence of intermediate goals leading to the final goal, and a low-level component is in charge of steering the agent from goal to goal. The main approaches to HRL are the options framework \citep{sutton1999between}, the feudal framework \citep{dayan1992feudal}, and the graph-based framework \citep{lee2022dhrl}. 


With {\bf options}, a different policy is used to achieve each of the intermediate and final goals, which are all considered terminal for their respective policy. In this framework, the option-critic algorithm \citep{bacon_option-critic_2017} addresses a {\em skill chaining} concern related to ours, but it does so with separate option policies. As we are interested in learning the low-level agent as a single policy, we do not cover the options framework further.

In all other approaches, the low-level agent is a GC policy which learns to achieve any local goal from its current state. During inference, the GC agent is iteratively conditioned on each intermediate goal until it reaches the final goal. Below, we focus on the way the low-level policy is learned.

In {\bf graph-based methods} \citep*{huang_mapping_2019,lee2022dhrl,kim_imitating_2023}, a GC policy is trained on single goals using \ddpg \citep{silver_deterministic_2014}, \tddd \citep{fujimoto_addressing_2018}, or \sac \citep{haarnoja_soft_2018}, combined with \her \citep{andrychowicz_hindsight_2018}. 

{\bf Feudal} approaches, such as \hac \citep{levy2019learning}, also use \ddpg+\her, while \hrac \citep{zhang_adjacency_2022} and \hiro \citep{nachum_data-efficient_2018} use dense rewards without \her. In contrast, \leap \citep{nasiriany_planning_2019} trains its low-level GC policy with \tdm \citep{pong2018temporal} where \ddpg is used to optimize a policy that is additionally conditioned on the remaining time to reach its goal. The reward function $r(s,g,t)=-I[t=T_{max}]d(s,g))$, motivates the agent to get as close to the goal as possible within the available time. 

In this work, we abstract away the differences between graph-based and feudal approaches by considering a generic planner that provides a sequence of intermediate goals given a final goal. In all these methods, the low-level policy treats each goal as if it were the only one without considering subsequent goals.
However, when goals are defined as a subset of the state space, they can be achieved in multiple configurations. In particular, when the agent reaches a goal that does not fully specify the corresponding state, nothing ensures that it is in a configuration that enables the next goal to be reached. This is the problem addressed in the next section.

\subsection{Following a single sequence of goals}

The \dcilii algorithm \citep{chenu_leveraging_2023} is an imitation learning algorithm designed to deal with the case in which chaining two subsequent goals can be an issue.
In \dcilii, the agent learns a behavior from a single demonstration that is split into a unique sequence of goals $\taug =
\{g_i\}_{i\in[0,N_{goals}]}$. The problem is defined as an extended MDP $M_{seq}=\{ S_{seq},A,R_{seq},P,\gamma,\mathcal{T},\taug \}$.
The agent observes $(s_t,g_t,i_t)\in S_{seq}= S\times G\times \mathbb{N}$, where $g_t\in G$ is the current goal the agent is targeting and $i_t$ the index of the goal in sequence \taug. When the agent takes an action, it moves to a new state, where $s_{t+1} \sim P(.|s_t,a_t)$ and $g_{t+1},i_{t+1}$ follows: 
$$g_{t+1},i_{t+1} = f_{DCIL}(s_{t},g_t,i_t)= \begin{cases}
    \taug[{i+1}],i+1 &\text{if } s_{t+1}\in S_{g_t}, \\
    g_{t},i_t  & \text{otherwise.} 
\end{cases} $$ 
As soon as the agent reaches a goal, it switches to the next one. The agent is rewarded each time it reaches its current goal with $R_{seq}(s_t,g_t,s_{t+1}) = \mathbb{1}[s_{t+1}\in S_{g_t}] $.\\
The low-level agent combines \sac+\her and maximizes the expected cumulative reward:
\begin{equation}
\Esp_{\substack{
    \\ s_0 \sim \rho_0
    \\ g_{t+1},i_{t+1} \sim f_{DCIL}(s_{t},g_t,i_t)
}}
\Bigg[\sum_{t=0}^{\infty} \gamma^t R_{seq}(s_t,g_t,s_{t+1}) \prod_{i=0}^{t}1-\mathbb{1}[s_i\in \mathcal{T}(\taug[N_{goals}])]\Bigg].
\end{equation}

The above formulation has two advantages. First, when the agent aims for its current goal, it also tries to reach it in a configuration that is valid for also reaching the rest of the sequence. Second, since the agent is also conditioned on the current goal index $i$, it can differentiate intermediate goals from terminal goals, which helps appropriately handling terminal conditions and using goal relabeling.

\section{Methods}

As the \dcilii agent learns from a single demonstration, the learned policy is only capable of imitating a single trajectory, not reaching an arbitrary goal. 
In this paper, we propose extending \dcilii to develop an agent capable of reaching any goal by following a sequence of intermediate goals.
We do so by conditioning the policy on the current goal and another goal, which can be either the next goal or the final one. But this raises another issue: When reaching one of its goals, should the policy consider it as terminal or not? In general, intermediate goals should not be considered terminal whereas for final goals, they can be terminal or not depending on the environment.

Thus, the approach we describe in the next section maintains two key properties: it prepares the agent for future goals and it properly handle terminal conditions.

\subsection{Following multiple sequences of goals}
\label{sec:FolMultSeqOfGoals}

We propose a method to solve a GC MDP $\mathcal{M}_{gc}$ (see Section~\ref{sec:gcrl}) by combining a low-level GC-agent and a high-level planner. Since our focus is on the low level, we utilize an expert planner that provides a sequence of intermediate goals from the agent's current state $s$ to a given final goal $fg$.
The low-level agent follows this sequence by iteratively reaching each intermediate goal up to the final one. The planner provides a function $next:S\times G \rightarrow G$, which determines the next intermediate goal the agent should reach based on its current state and its final goal.

As in \dcilii, we encourage the agent to prepare for future goals while pursuing intermediate ones by rewarding each goal reached and integrating successive goals into the transition function. However, if we were to rely on the GC-MDP $\mathcal{M}_{gc}$, where the agent is conditioned only on its current state and goal, the transition $(s_t, bg_t) \rightarrow (s_{t+1}, bg_{t+1})$ would contain a transition between goals that depends on an unobserved final goal, making the process a Partially Observable Markov Decision Process. In \dcilii, the problem of non-Markovian goal transitions was resolved by adding the ID of the current goal to the observation. In the multi-goal case, ensuring that goal transitions remain Markovian requires including the final goal in the observation: $(s_t, bg_t, fg) \rightarrow (s_{t+1}, bg_{t+1}, fg)$. In this setup, the next goal $bg_{t+1}$ can be implicitly determined from the observation, as shown in \eqref{eq:seq-goal-dynamic}.

In order to properly formalize this process, we propose the following MDP formulation:
$\mathcal{M}_{gseq}=(S_{gseq},A,R_{gseq},P,\gamma,\mathcal{T})$.
At the beginning of the episode, the planner computes the sequence of intermediate goals up to the final goal $fg$. At each step, the agent observes $(s_t,bg_t,fg)\in S_{gseq}=S\times G \times G$ which is composed of the state $s_t$, a behavioral goal $bg_t$ that the agent must reach, and the final goal $fg$ which stays fixed during the episode. When the agent takes an action, it moves to a new state, where $s_{t+1} \sim P(.|s_t,a_t)$ and $bg_{t+1}$ follows: 
\begin{equation}\label{eq:seq-goal-dynamic}
bg_{t+1} = f_{gseq}(s_t,bg_t,fg) =\begin{cases}
    next(s_{t+1},fg)&\text{if } s_{t+1}\in S_{bg_t}, \\ 
    bg_t  &\text{otherwise.}
\end{cases}
\end{equation}
The reward function is defined as $R_{gseq}(s_t,bg_t,s_{t+1}) =  \mathbf{1}[s_{t+1}\in S_{bg_t}]$ so that the agent is rewarded for each behavioral goal reached during the trajectory. Finally, $\mathcal{T}(fg)$ is the set of terminal state according to the current goal $fg$. 
The objective is to find a policy that maximizes the classic expected cumulative reward:
\begin{equation}
\Esp_{
    \substack{
    \\ s_0 \sim \rho_0
    \\ fg \sim P(fg)
    \\ bg_{t+1} \sim f_{gseq}(s_t,bg_t,fg) 
    }}
\Bigg[\sum_{t=0}^{\infty} \gamma^t R_{gseq}(s_t,bg_t,s_{t+1}) \prod_{i=0}^{t}1-\mathbb{1}[s_i\in \mathcal{T}(fg)]\Bigg].
\end{equation}

Based on this formulation, we can directly use \tddd to learn a policy.
The advantages of this formulation are the following. First, the agent is rewarded for chaining all goals of the trajectory, not only for reaching its current goal.
Second, it enables proper handling of terminal states, as we can differentiate intermediate and final goals.

$\mathcal{M}_{gseq}$\textbf{-Relabeling}: Since the agent is conditioned on two goals, each of them can be relabeled.
Given a sample $(s_t,bg_t,fg,s_{t+1},bg_{t+1})$, we can relabel $bg_t \rightarrow ag_{k_1}$ and $fg\rightarrow ag_{k_2}$ where $ag_t$ represents the achieved goal at step $t$ and $t<k1<k_2<t_{max}$.\\
However, $\mathcal{M}_{gseq}$ was made to also integrate successive goals in the transition dynamic: Once the current goal is reached, the next goal must be updated to be the one the planner would have selected next to reach the final goal. When $bg_t$ and $fg$ are relabeled, $bg_{t+1}$ is updated according to \eqref{eq:seq-goal-dynamic}. 
Skipping this step would cause the agent to learn from transitions between goals that violate the Markov property. However, this re-planning step for each relabeled transitions can be prohibitively costly if paths cannot be precomputed and cached.

The full pseudocode for this algorithm is provided in Appendix~\ref{app:pseudo code}.
In the next section, we introduce a computationally more efficient method that does not require planning during relabeling.

\subsection{Two-goal sequence following: formalization}\label{sec:Two-goal}

In feudal methods, the low-level agents only optimize the way they reach the next goal without taking the sequence into account. In Section~\ref{sec:FolMultSeqOfGoals}, we proposed a method to optimize the way the agent reaches all goals in the sequence. Now we propose an intermediate formulation: the agent only optimizes the way it reaches a set of two goals. 
Despite the fact that the agent ignores the planner during its policy and critic updates, it can still be used to traverse any sequence of goals, by iteratively conditioning the policy on the upcoming two goals within a planner-provided sequence of goals.

We use the following MDP to formalize this objective: $M_{2G} = \{ S_{2G},A,R_{2G}, P,\gamma, \mathcal{T}\}$. 
At each step, the agent observes $(s_t,bg_t,fg)\in S_{2G}=S\times G\times G$ where $s$ is the current state of the agent, $bg_t$ is the first goal that the agent must reach and $fg$ is the second. In practice, those goals correspond to the first two goals of a sequence provided by the planner.
When the agent takes an action, it moves to a new state $s_{t+1}\sim P(.|s_t,a_t)$, the final goal $fg_t$ stays fixed and $bg_{t+1}$ follows: 
\begin{equation}\label{eq:2G-goal-switching}
    bg_{t+1} = f_{2G}(s_t,bg_t,fg)= \begin{cases}
    fg &\text{if } s_{t+1}\in S_{bg_t},\\
    bg_t & \text{otherwise.}
\end{cases}
\end{equation}

The agent is rewarded for reaching $bg_t$ though $R_{2G}(s_t,bg_t,s_{t+1}) = \mathbb{1}[s_{t+1}\in S_{bg_t}]$. 
In this simplified goal-switching dynamic, the agent must learn how to always be able to reach a first goal in a configuration compatible with reaching the second one.
Finally, $\mathcal{T}(fg)$ is the set of terminal state according to the final goal $fg$. 
This formulation being a special case of an MDP, the objective is still to find a policy that maximizes the expected cumulative reward:
\begin{equation}
\Esp_{
    \substack{
    \\ s_0 \sim \rho_0
    \\ fg \sim P(fg)
    \\ bg_{t+1} \sim f_{2G}(s_t,bg_t,fg)
    }}
\Bigg[\sum_{t=0}^{\infty} \gamma^t R_{2G}(s_t,bg_t,s_{t+1}) \prod_{i=0}^{t}1-\mathbb{1}[s_i\in \mathcal{T}(fg)]\Bigg].
\end{equation}

$\mathbf{M_{2g}}$\textbf{-Relabeling}
Given a sample $(s_t,bg_t,fg,s_{t+1},bg_{t+1})$, we can relabel $bg_t \rightarrow ag_{k_1}$ and  $fg_t\rightarrow ag_{k_2}$ where $ag_t$ represents the goal achieved at state time step $t$ and $t<k1<k_2<t_{max}$.\\
Similarly to the previous example, we must ensure that the goal switching mechanism is valid within $M_{2GC}$, so that $bg_{t+1}$ is updated following \eqref{eq:2G-goal-switching}.
This method requires less computation during the update steps while still being less myopic than an agent conditioned on a single goal. However, in complex cases, only preparing for the two next goals may be insufficient.

The full pseudocode for this algorithm is provided in Appendix~\ref{app:pseudo code}.

\section{Experiments}

Our experimental study is designed to compare agents that prepare for the next or the final goals together with myopic agents that do not prepare at all and non-sequential agents which ignore intermediate goals. In particular, we define three low-dimensional environments with various goal chaining difficulties and various terminal conditions to highlight the pros and cons of the different approaches. We compare their success rate and time-to-goal, as considering an intermediate goal as terminal or not can have a large effect on the way the agent reaches goals.

\subsection{Environments}
\begin{figure}[h]
    \centering
    \includegraphics[width=0.70\linewidth]{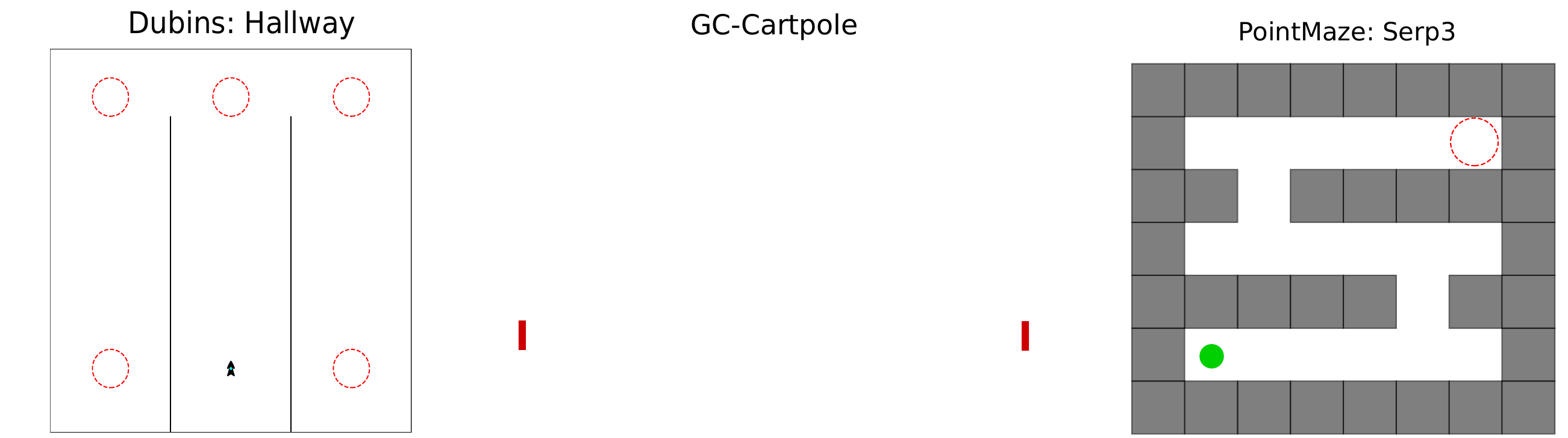}
    \caption{The environments used in our experiments. In \dubins, the agent is evaluated on five goals, starting in the main corridor and aiming for each goal shown as a red circle. In \cartpole, the agent is evaluated on the two goals farthest from the center, shown as red rectangles. In \pointmaze, the agent, shown as a green ball, is evaluated only on the hardest goal.}
    \label{fig:environments}
\end{figure}

\textbf{Dubins Hallway} is a navigation task where the agent controls a car in a 2D maze. The state $s=\{x,y,cos(\theta),sin(\theta)\}$ includes the position and orientation of the agent. Each goal $g$ is defined as a position $(x,y)$ and is considered reached if the agent is within a ball of radius $0.1$ centered at that position. The agent moves forward at a fixed speed at each step, the action controls the variation of the orientation $\dot{\theta}$ for the agent.
During a training episode, both the agent's initial position and the goal are randomly sampled within the maze boundaries. If the agent hits a wall, it stays stuck until the episode ends. The state is only terminal when the agent reaches the goal. A key feature of this environment is that, when the agent is in the central hallway, it should not prepare for the next goals in the same way depending on whether the final goal is on the left- or right-hand side. In contrast, the velocity of the agent being constant, the time-to-goal does not vary much in this environment, so we do not present results on this aspect.

In \textbf{Goal-Conditioned Cartpole}, the agent must reach a given position while balancing a pole. The state $s=(\dot{x},\theta,\dot{\theta})$ contains velocity $\dot{x}$, angle $\theta$, and angular velocity $\dot{\theta}$. As explained in more details in Appendix~\ref{app:rel_goals}, the agent's goal is the difference between its current position $x$ and a fixed target position $x_{dg}$. The goal is reached once $||x-x_{dg}||<0.05$. Actions are continuous values in $[-1,1]$ proportional to the force applied to the cart. During a training episode, the agent targets a random uniform goal in the range $[-5,5]$. A state is terminal when $\theta$ leaves the $[-0.12^{\circ},0.12^{\circ}]$ interval. As reaching a goal is not terminal, the agent must learn to reach its target position and stay there while keeping the pole balanced.
 
\textbf{PointMaze Serp3} \citep{gymnasium_robotics2023github} is a navigation task where the agent controls a ball in a 2D maze. The state $s=\{x,y,\dot{x},\dot{y}\}$ includes the position and velocity of the agent. Each goal $g$ is defined as a $(x,y)$ position and is considered reached if the agent is within a ball of radius $0.45$ centered at that position. The action represents the linear force exerted on the ball in the x and y directions. In \citep{gymnasium_robotics2023github}, the agent's velocity was capped; we have removed this constraint. During a training episode, both the agent's initial position and the goal are randomly sampled within the maze boundaries. The state is only terminal when the agent reaches the goal. As shown in \figurename~\ref{fig:environments}, the agent is only evaluated on the farthest goal.

\subsection{Compared agents}

We compare the following methods:  
\par$\bullet$ \tdddher: An agent only conditioned on the final goal,
\par$\bullet$ \myopictddd: An agent conditioned iteratively on the successive goals provided by the expert planner. This is the formulation adopted by most feudal and graph-based approaches (see Section~\ref{sec:feudal}),
\par$\bullet$ \mseqtddd : the agent presented in Section~\ref{sec:FolMultSeqOfGoals},
\par$\bullet$ \twogtddd: the agent presented in Section~\ref{sec:Two-goal}.
\par
All these agents rely on \tdddher with the same hyper-parameters (see Appendix~\ref{app:hyper}).

\subsection{Results in Dubins Hallway}

\begin{figure}[htbp]
    \centering
    \includegraphics[width=\linewidth]{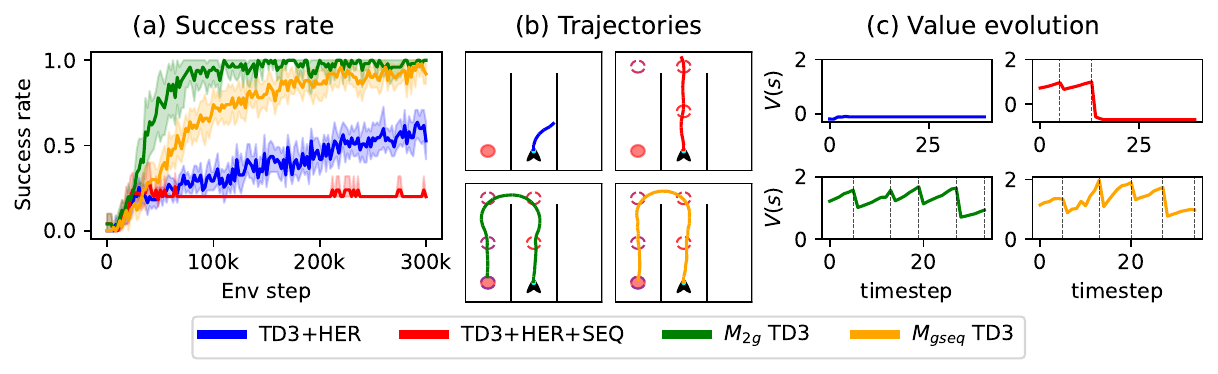}
    \caption{\textbf{(a)} Comparison of Success Rate for each methods. Evaluation is performed 10 times every 2K steps, with results reported as the mean 95\% confidence interval over 10 seeds.
\textbf{(b)}: Set of trajectories of trained agents on \dubins.  
\textbf{(c)}: Value function over episodes matching trajectories in (b). Our sequential approaches outperform the myopic and non-sequential agents. The \mseqtddd agent struggles to propagate value over full episodes.}
    \label{fig:DubinsPerf}
\end{figure}

In this section, we evaluate all methods in \dubins.

Figure~\ref{fig:DubinsPerf}(a) presents the mean success ratio across the five goals depicted in \figurename~\ref{fig:environments}(a). \myopictddd plateaus at 20\%, indicating that it has failed to achieve any challenging goal. \tdddher does not fully converge, but it manages to reach slightly more than half of the evaluation goals. In contrast, our two formulations, \twogtddd and\mseqtddd, successfully reach all goals.

To further analyze the behavior of each algorithm, \figurename~\ref{fig:DubinsPerf}(b,c) presents a representative trajectory and the corresponding value function over time for each method. The value function is estimated as $V(s_t)\simeq Q(s_t,\pi(a|s))$.
In the selected run, \tdddher exhibits an almost flat value function, indicating that it has failed to back-propagate rewards effectively.  \myopictddd focuses solely on its next goal, and its value increases toward 1 when the goal appears attainable. However, when it reaches the second goal in a configuration where it is too late to turn, the value drops sharply to 0.
The value profile of \twogtddd is often above 1, suggesting that it evaluates the policy as capable of reaching both its current and next goals. At the end of the trajectory, its value reaches 1, which corresponds to the final reward before transitioning to a terminal state.
\mseqtddd also manages to turn; however, its value profile does not align with the optimal value function of $\mathcal{M}_{gseq}$. An optimal agent would receive rewards for all future goals, leading to a staircase-like decrease in the value function. Instead, we observe that value updates have only propagated from the next two goals or, at best, the next three goals, which suggests a weakness of \mseqtddd in front of long range value propagation.

\subsection{Results in Cartpole \& PointMaze}

\begin{figure}[!htbp]
    \centering
    \includegraphics[width=\linewidth]{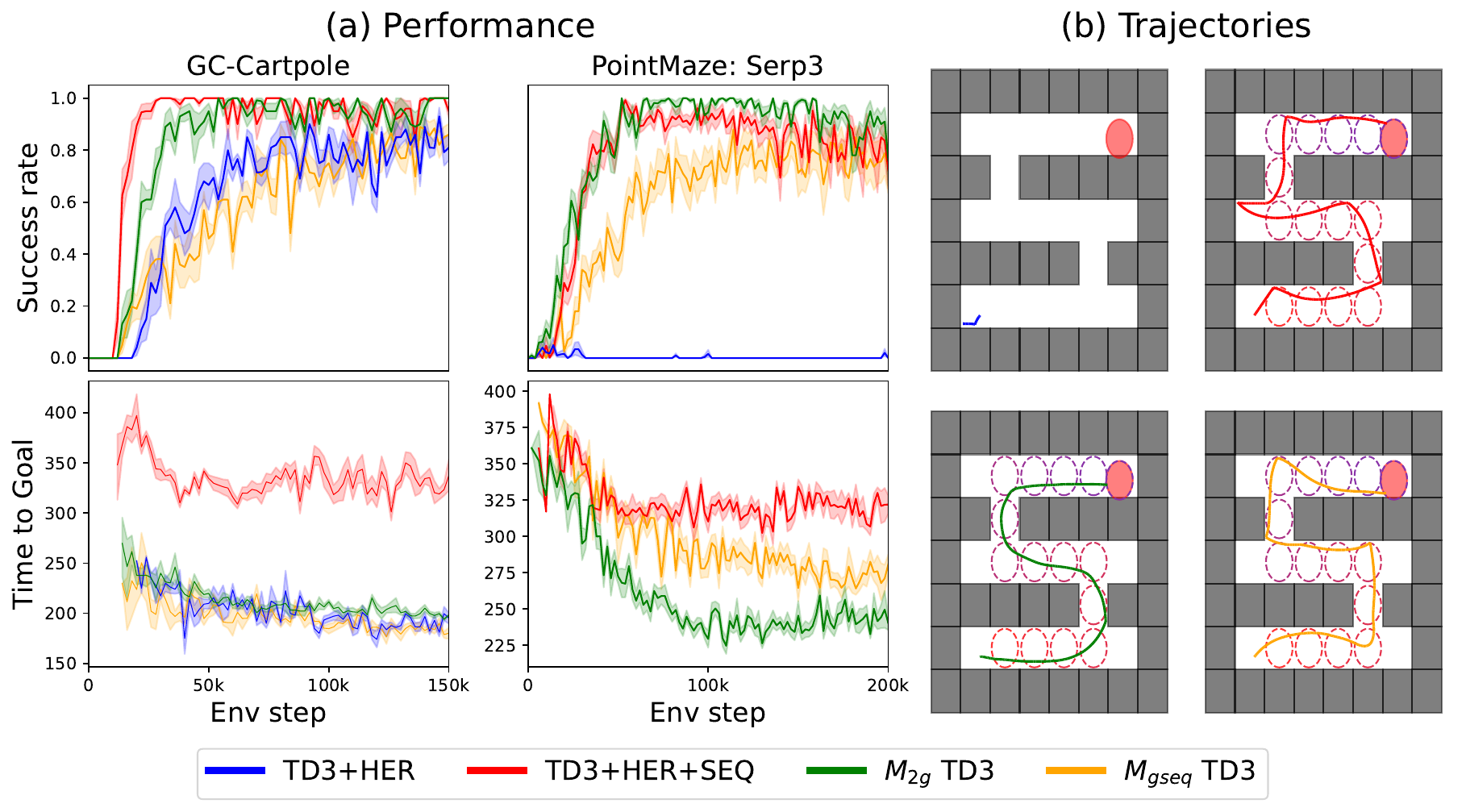}
    \caption{\textbf{(a) Top:} Comparison of success rates for each method. Evaluation is performed 10 times every 2K steps, with results reported as the mean and 95\% confidence interval over 10 runs. For all runs, metrics are smoothed using a moving average with a window size of 5 to increase readability.
    \textbf{(a) Bottom:} Time required to reach the goal state, considering only successful trajectories for computing the mean and confidence interval.
    \textbf{(b)} A set of trajectories generated by trained agents on \pointmaze. Again, \twogtddd seems to outperform \mseqtddd.}
    \label{fig:Cartpole_PointMaze_Perf}
\end{figure}

In \cartpole, all methods successfully achieve the evaluation goals in most cases (see \figurename~\ref{fig:Cartpole_PointMaze_Perf}(a)). However, \myopictddd takes longer to reach the evaluation goal. As the state is never terminal for reaching a goal, the optimal behavior of \myopictddd decelerates before reaching each goal, to ensure it could stop at equilibrium there and receive a reward of $+1$ for all subsequent steps. In contrast, the optimization process for both \twogtddd and \mseqtddd is structured such that reaching an intermediate goal yields only a $+1$ reward and triggers a subsequent goal transition. As a result, these methods favors passing through each intermediate goal without slowing down, enabling them to achieve a speed similar to \tddd, which directly targets the final goal.

In \pointmaze, iteratively conditioning the agent on intermediate goals is very efficient, as shown in \figurename~\ref{fig:Cartpole_PointMaze_Perf}(a). In particular, \tddd, which is only conditioned on the final goal, fails to reach distant goals. \myopictddd completes long sequence of goals, but the agent does so in a suboptimal way, moving too quickly and bouncing off walls as shown in \figurename~\ref{fig:Cartpole_PointMaze_Perf}(b). This behavior results from its MDP formulation, in which the agent always tries to reach the next goal as quickly as possible. \twogtddd, on the other hand, prepares for the next two goals and achieves both the fastest speed and stability. \mseqtddd learns at a slower rate, and does not reach the same success rate as the other methods following sequence of goals. Again, slow value propagation might be the issue.

\subsection{Ablations}

Across all previously used environments, we assess the importance of relabeling both the current and final goals in \twogtddd and \mseqtddd.

\begin{figure}[!htpb]
    \centering
    \includegraphics[width=0.75\linewidth]{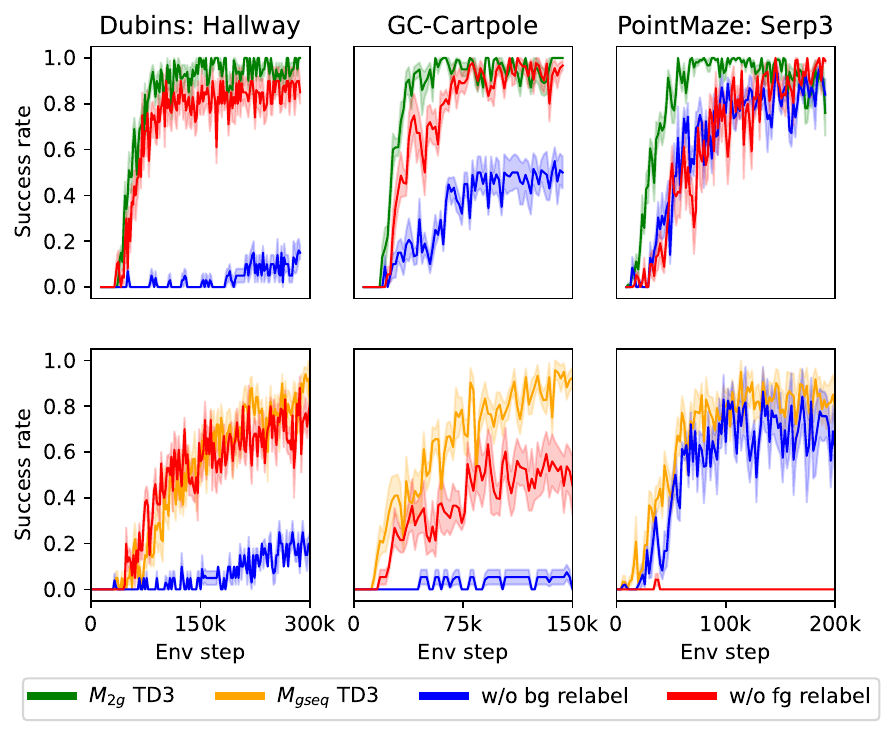}
    \caption{Ablation study: Success rate for all previous environments. Evaluation is performed 10 times every 2K steps, with results reported as the mean and 95\% confidence interval over 10 runs. For all runs, metrics are smoothed using a moving average with a window size of 5 to increase readability. \textbf{Top:} Ablation study for \twogtddd, where the blue and red curves represent the removal of the first and second relabeling mechanisms, respectively. \textbf{Bottom:} Ablation study for \mseqtddd, where the blue and red curves correspond to the removal of the first and final goal relabeling mechanisms, respectively.}
    \label{fig:ablation}
\end{figure}

As shown in \figurename~\ref{fig:ablation}, in \pointmaze, omitting to relabel the current goal has minimal impact. This is because the graph guiding the agent is dense (as shown in \figurename~\ref{fig:expert_graph}) and the threshold required to reach a goal is large, thus the task reward cannot be considered sparse. However, in all other environments, excluding current goal relabeling significantly affects performance.
In \twogtddd, while omitting final goal relabeling also impacts performance, its effect is less pronounced. For \mseqtddd omitting the final goal relabeling substantially reduces performances for \cartpole and \pointmaze.

\subsection{General discussion}
 
Our experimental results indicate that, beyond outperforming the myopic and the non-sequential agents, \twogtddd performs better than \mseqtddd.
Indeed, agents trained on $M_{2G}$ learned faster and with greater stability than agents trained on $M_{seq}$, which struggle to learn an optimal policy. Besides, \mseqtddd is more sensitive to the ablation of \her. A first explanation for this better performance of \twogtddd could be that $M_{2g}$ is defined on a shorter temporal horizon, facilitating value propagation.

Furthermore, while both \twogtddd and \mseqtddd condition on a single state and two goals, the \mseqtddd variant faces much higher diversity because its final goal can be any goal in the goal space. In contrast, in \twogtddd, the final goal corresponds to the next goal on the path provided by the planner. This high diversity forces \mseqtddd to generalize over a wider set of inputs, which makes learning more difficult.

However, \twogtddd benefits from a less global information. As a consequence, its performance may degrade in more complex environments. In such cases, an agent might reach an initial goal in a valid configuration to reach the immediate next goal, yet that configuration may not be compatible with reaching subsequent goals. We keep the evaluation of \twogtddd and \mseqtddd in more difficult environments for future work.

\section{Conclusion}
In this paper, we have addressed the problem of following diverse sequences of low-dimensional goals when a planner is provided. We have shown that navigating goal sequences by considering only the next goal can lead to failure cases. We proposed the $M_{seq}$ MDP framework, in which we formalize goal transitions and terminal states to ensure the objective is to reach all goals in the sequence. Additionally, we proposed the $M_{2g}$ MDP, where the objective is solely to reach the next two goals in the sequence. Through navigation and pole-balancing experiments, we have shown that agents trained with $M_{2g}$ were more stable and sample-efficient. One of the main limitations of this work is the use of a fixed planner. While it is possible to access one through expert knowledge, this element is often learned. Since successive goals are integrated into the transition function of our MDP, having a planner that evolves over time would cause the low-level MDP on which the GC agents are trained to change as well, making the learning process more challenging.

\subsubsection*{Acknowledgments}
\label{sec:ack}



\bibliography{main}
\bibliographystyle{rlj}

\appendix
\beginSupplementaryMaterials

\label{sec:appendix1}

\section{Pseudo-code}
\label{app:pseudo code}

Algorithm \ref{alg:main} provides the pseudocode for the main loop, illustrating how \twogtddd and \mseqtddd operate within the environment using a planner. Additionally, Algorithm \ref{alg:learnstep} outlines the relabeling mechanism for both methods.

\begin{algorithm}[htpb]
\caption{Main Loop}\label{alg:main}
\begin{algorithmic}
\Require{$Q$,$\pi$,planner,env}
\State $RB \gets []$
\For{$N = 1 : N_{episodes}$}
    \State  $s,env_g \gets env.reset()$ 
    \State $path \gets [g_1,g_2,...,env_g] \gets planner.path(s,env_g)$
    \State  $bg \gets path[0]$
    \If{$algo =\mathcal{M}_{2g}$}
        \State $fg \gets path[1]$
    \EndIf
     \If{$algo =\mathcal{M}_{gseq}$}
        \State $fg \gets env_g$
    \EndIf
    \State $done \gets false$
    \While {not done}
    
    \State $a \gets\pi(a|s,bg,fg)$
    \State $s',r,term,trunc \gets \text{env.step}(a)$

    \If{$s' \in bg$}  
        \State $path \gets [g_1,g_2,...,env_g] \gets planner.path(s,env_g)$
        \State  $bg' \gets path[0]$
        \If{$algo =\mathcal{M}_{2g}$}
            \State $fg \gets path[1]$
        \EndIf
    \EndIf
    \State $RB \gets RB + (s,a,r,bg,fg,s',bg',term)$ 

    \State $s,bg = s',bg'$
    
    \State $Q,\pi \gets  \text{Learn\_Step}(\pi,Q,RB,\text{next},\text{env.terminal\_func})$ 
    \State $done \gets term \vee trunc$
    \EndWhile
\EndFor
\end{algorithmic}
\end{algorithm}

\begin{algorithm}[htpb]
\begin{algorithmic}
\caption{Learn\_Step}\label{alg:learnstep}
    \Require{Q,$\pi$,RB,planner,terminal\_func}
    \State $\tau \gets \text{Sample trajectory from RB}$ 
    \State $ag[] \gets \tau$ \Comment{Get all achieved goal in the trajectory}
 \State $s_t,a_t,bg_t,r_t,fg,s_{t+1},bg_{t+1},term_t \gets \text{Sample from $\tau$}$
 \State $relabel \gets \text{(random($0.0$, $1.0$) $<$ 0.8)}$
 \If{$relabel$}  
        \State $k_2 \sim random(t,t_{max})$
        \State  $fg \gets ag_{k_2}$ 
        \State $k_1 \sim random(t,k_2)$
        \State  $bg_t \gets ag_{k_1}$ 
 \EndIf
 \If{$s_{t+1} \in bg_t$}  
        \If {$algo = \mathcal{M}_{2g}$}
            $bg_{t+1} \gets fg$ 
        \EndIf
        \If {$algo = \mathcal{M}_{gseq}$}
            $bg_{t+1} \gets \text{planner.next}(s_t,fg)$ 
        \EndIf
 \EndIf
 \State $r_t \gets 1$ \textbf{if}  $s_{t+1} \in bg_t$ \textbf{else} $0$

 \State $term_t \gets \text{terminal\_func}(s_{t},a_{t},s_{t+1},fg)$
 \State $\text{transition} = s_t,a_t,bg_t,r_t,s_{t+1},bg_{t+1},fg,term_t$
  \State $\pi, Q \gets \text{TD3\_Update(transition,$\pi$,$Q$ )}$
\end{algorithmic}
\end{algorithm}

\section{Expert graph}
\label{app:expert_graph}

\begin{figure}[H]
    \centering
    \includegraphics[width=0.6\linewidth]{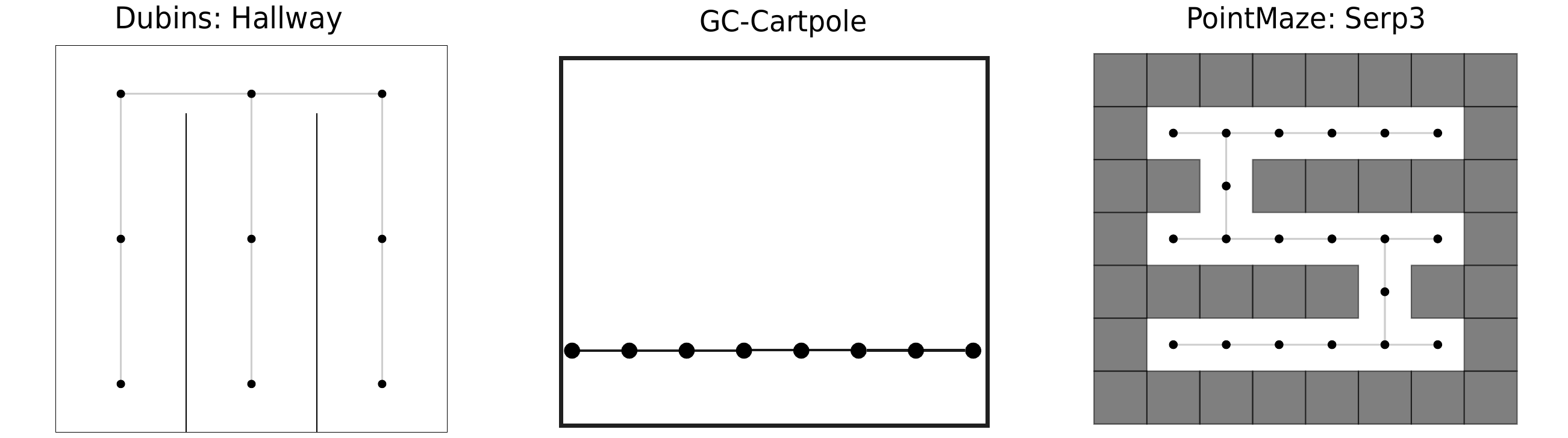}
    \caption{ For each environment, graph used to implement the expert planner.}
    \label{fig:expert_graph}
\end{figure}

The expert planner finds the shortest path using handcrafted graphs shown in \figurename~\ref{fig:expert_graph}.

\section{Absolute and relative goals}
\label{app:rel_goals}
Absolute goals are goals that correspond to absolute positions in the goal space, while relative goals are goals whose positions are relative to the agent. 

The relative goal formulation is better suited in \cartpole. Indeed, with an absolute goal formulation, going from $x=-2$  to $x=0$ and from $x=8$ to $x=10$ are two different things, whereas with a relative goal formulation, the agent must only learn how to go to $x_{rel}=2$ to solve both cases at once.

\section{Hyper-parameters}
\label{app:hyper}

The hyper-parameters used for each environment are presented in Table~\ref{tab:hyper_parameters}. 

\begin{table*}[ht!]
\centering
\caption{Hyper-parameters used for TD3}
\begin{tabular}{ |p{2.9cm}|p{1.5cm}|p{1.9cm} |p{1.5cm}| }
 \hline
 Hyper-parameters & \dubins & \pointmaze & \cartpole \\
 \hline
  Random actions  & $5K$ & $5K$ & $5K$ \\
 Critic hidden size  & $[256,256]$ & $[256,256]$ & $[256,256]$ 
 \\
 Policy hidden size & $[256,256]$ & $[256,256]$ & $[256,256]$ 
 \\
 Activation functions & ReLU & ReLU & ReLU
 \\
 Batch size & 256 & 256 & 256
 \\
 Discount factor & 0.95 & 0.99 & 0.99
 \\
 Critic lr & $1\times10^{-3}$ & $1\times10^{-3}$ &  $1\times10^{-3}$
 \\
 Policy lr & $1\times10^{-3}$ & $1\times10^{-4}$ & $1\times10^{-4}$ 
 \\
 HER Relabel  & 80\%  & 80\% & 80\%  \\
 \hline
\end{tabular}
\label{tab:hyper_parameters}
\end{table*}


\end{document}